\documentclass[conference]{IEEEtran}
\IEEEoverridecommandlockouts
\usepackage{cite}
\usepackage{amsmath,amssymb,amsfonts}
\usepackage{algorithmic}
\usepackage{graphicx}
\usepackage{textcomp}
\usepackage{xcolor}
\def\BibTeX{{\rm B\kern-.05em{\sc i\kern-.025em b}\kern-.08em
    T\kern-.1667em\lower.7ex\hbox{E}\kern-.125emX}}
\begin{document}

\title{Syntactic Evolution in Language Usage\\}

\author{\IEEEauthorblockN{Surbhit Kumar}
\IEEEauthorblockA{\textit{Department of Computer Science} \\
\textit{Rochester Institute of Technology}\\
Rochester, New York \\
sk2768@rit.edu}
}

\maketitle

\begin{abstract}
This research aims to investigate the dynamic nature of linguistic style throughout various stages of life, from post teenage to old age. By employing linguistic analysis tools and methodologies, the study will delve into the intricacies of how individuals adapt and modify their language use over time. The research uses a data set of blogs from blogger.com from 2004 and focuses on English for syntactic analysis. The findings of this research can have implications for linguistics, psychology, and communication studies, shedding light on the intricate relationship between age and language.
\end{abstract}


\section{Introduction}
Language is one of the most ancient knowledge that humans have evolved. From the time of homo-sapiens when there might have been no or little words with voice, a gift that would have been naturally acquired, seems to date back about 300,000 years as per the research. However, other civil development took way longer and only dates back to 40,000 years.\cite{b10} What I intend to do here is to study a much shorter duration and see if languages do change their shape and texture within a lifespan and a language. Hence for this study, I have majorly focussed on English, across a lifespan of three broad categories of age groups Young (18-34 years old), middle-aged (35-41 years old), and Old (42 years and older) to evenly balance the dataset used for the analysis after pre-processing. 

The dataset that I have used comes from the publically available source of blogger.com which was one of the few social media platforms available from the 1990s to the early 200s.\cite{b11} This dataset dates back between 2002-04 and contains text information with the author's age when the blog text and the comment were written. I have derived the age group based on that so that the evolution can be broadly studied. There are about 450,000 rows of text in the dataset, which were finally reduced to about 52,000 rows after pre-processing to balance data. While the demographic details of these users (such as geographic location, and interests) were not explicitly available, it is reasonable to infer that they likely represent a tech-savvy, internet-engaged demographic, potentially skewing towards younger age groups who were more inclined to participate in online blogging at that time.

The following example sentences from the dataset, share different age groups and different complexities\\
Young: "Love pictures, baby!"\\
Middle-Aged: "Love those pictures of Tim Peretti."\\
Old: "I actually love to teach about the first Apostles, especially since my mother gave me a set of paintings done by an artist, a woman, who wanted to paint what she beleived the Apostles would have looked like."

In principle, the hypothesis is that with time language learning becomes more evolved and we learn to use and understand more complex language structures. Hence the study will involve syntactic feature analysis of the different age groups and also will attempt to forecast a group based on text to see if the study can be used in the future for the cases where communication may need modulation based on the age group of the person, like education.

\section{Research Design}
\subsection*{Design of experiment}
\begin{figure}[htbp]
    \centering
    \includegraphics[width=1\linewidth]{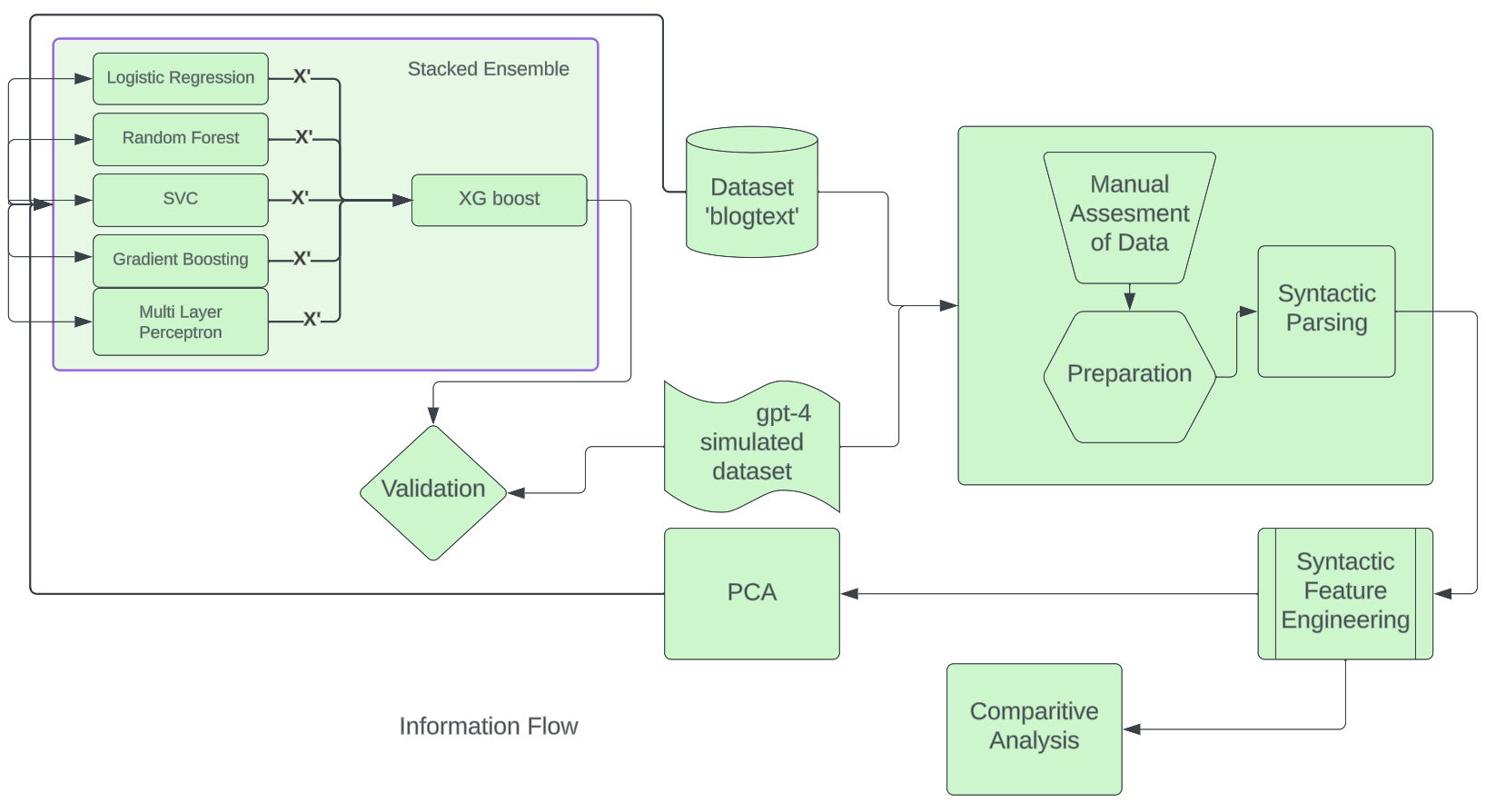}
    \caption{Information Flow Diagram}
    \label{fig:arch}
\end{figure}

The overall design of the experiment involved multiple steps which can be briefly illustrated in the following diagram and the description that follows.(fig. \ref{fig:arch})
High-level Steps:\\
1. Performing raw data analysis, manual and WordCloud\\
2. Preprocessing the data (removal of $<$18yr) and balancing dataset for forecasting\\
3. Feature engineering using syntactic parsing\\
4. Syntactic feature comparison of real user data from bloggers.com with GPT4 output text, using openai API\\
5. Forecasting age\_group using 5-component PCA on syntactic features and 2-layered multi-model stacked ensemble \cite{b12}\cite{b13}
\subsection*{Syntactic Parsing}
\begin{figure}[htbp]
    \centering
    \includegraphics[width=0.5\linewidth]{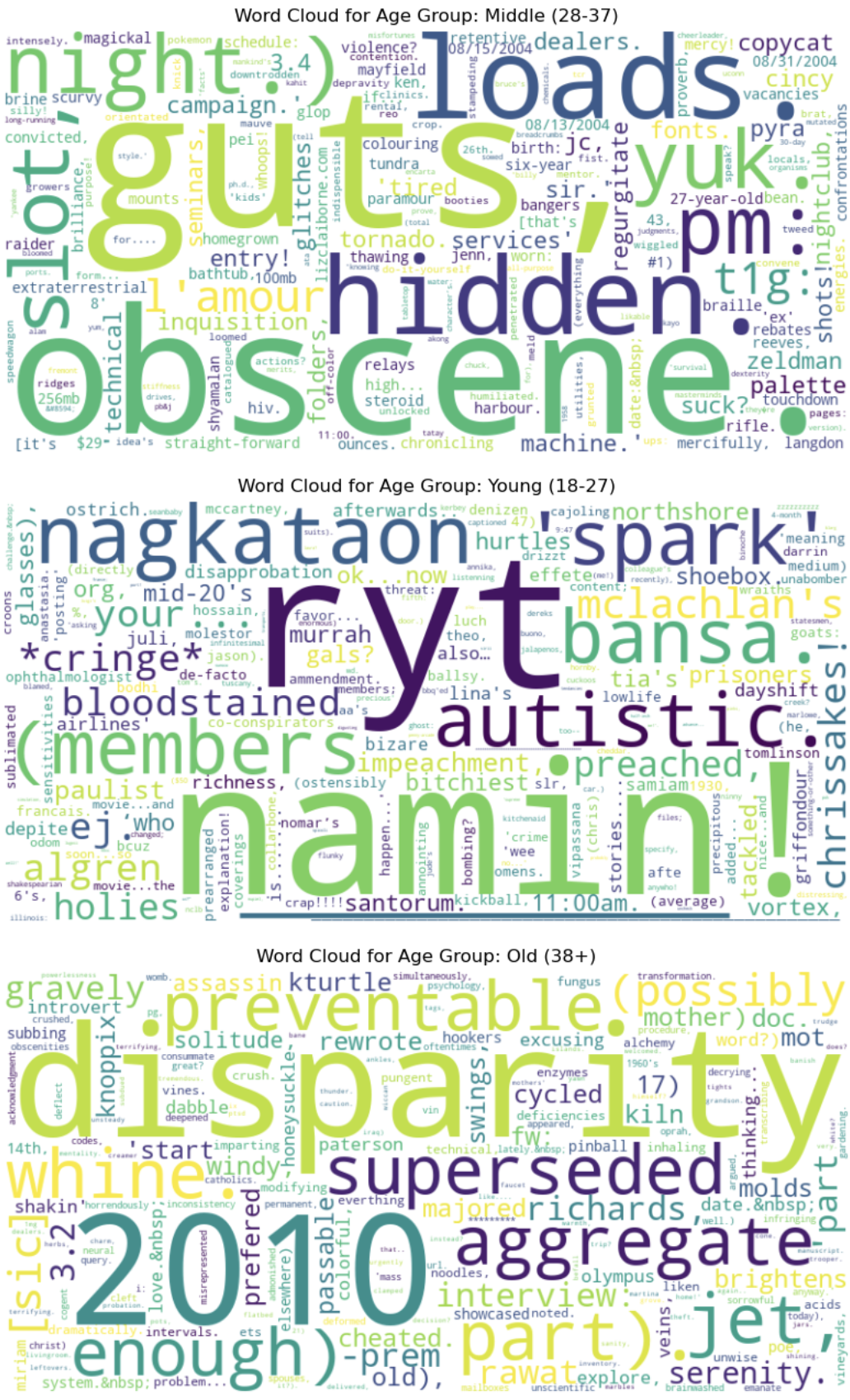}
    \caption{WordCloud of BlogText dataset by age group}
    \label{fig:blogtextWC}
\end{figure}

Word Cloud Analysis:

Before immersing myself in syntactic analysis, let us see a word cloud to illustrate high-frequency words in different age groups. This method likely granted me an intuitive grasp of the most prevalent words associated with each age group. (fig. \ref{fig:blogtextWC})

\subsubsection*{Analysis of Syntactic Features}
The study encompassed a comprehensive analysis of syntactic elements, including rates and ratios of various parts of speech and syntactic structures. These elements comprise 'Noun Rate', 'Verb Rate', 'Demonstrative Rate', 'Adjective Rate', 'Pronoun Rate', 'Adverb Rate', 'Conjunction Rate', 'Possessive Rate', 'Noun-Verb Ratio', 'Noun Ratio', 'Pronoun-Noun Ratio', 'Closed-class Word Rate', 'Open-class Word Rate', 'Content Density', 'Idea Density', 'Proportion of Inflected Verbs', 'Proportion of Auxiliary Verbs', 'Proportion of Gerund Verbs', 'Proportion of Participles', 'Number of Clauses', and various others \cite{b1} \cite{b2}. These metrics were meticulously calculated for each textual entry and subsequently aggregated to derive mean values for overarching analysis, though individual row calculations may be subject to variance-induced noise.

\textit{Elucidation of Key Terms:}
\begin{itemize}
    \item Rates such as 'Adjective Rate' and 'Noun Rate' indicate the percentage of text composed of adjectives and nouns, respectively. This approach is similarly applied to other part-of-speech rates.
    \item Syntactic characteristics including 'Clause Rate', 'Yngve Depth', and 'Discourse Marker Rate' provide insights into sentence structure and the distribution of specific word types, like pronouns and gerunds \cite{b4}.
    \item Metrics like 'Idea Density' and 'Mean Yngve Depth' delve into the complexity and structural depth of syntactic constructs within the text.
\end{itemize}

\textit{Divergence Across Age Groups:}
The syntactic analysis revealed notable variations across different age groups in aspects such as word type usage rates, content density, and overall sentence structure.

\textit{Interpreting Specific Syntactic Features:}
\begin{itemize}
    \item 'References to Self Rate': Indicates the frequency of self-references within the text.
    \item 'Unique Words Rate': Represents the diversity of vocabulary, indicated by the proportion of unique words used.
\end{itemize}
\cite{b6}\cite{b7}\cite{b8}

\subsection*{Generating text from GPT-4}
A Python script was used that integrated with the OpenAI API to generate text based on different age groups and topics. It began by importing necessary libraries and setting up the OpenAI client with an API key. It used OpenAI API to generate a piece of text. It constructs a prompt that includes the age group and a specified topic and then calls the OpenAI API to get a text completion. The generated text is intended to be a short sample of up to 20 words, reflective of the given age group and topic.\cite{b3}
\begin{figure}[htbp]
    \centering
    \includegraphics[width=0.5\linewidth]{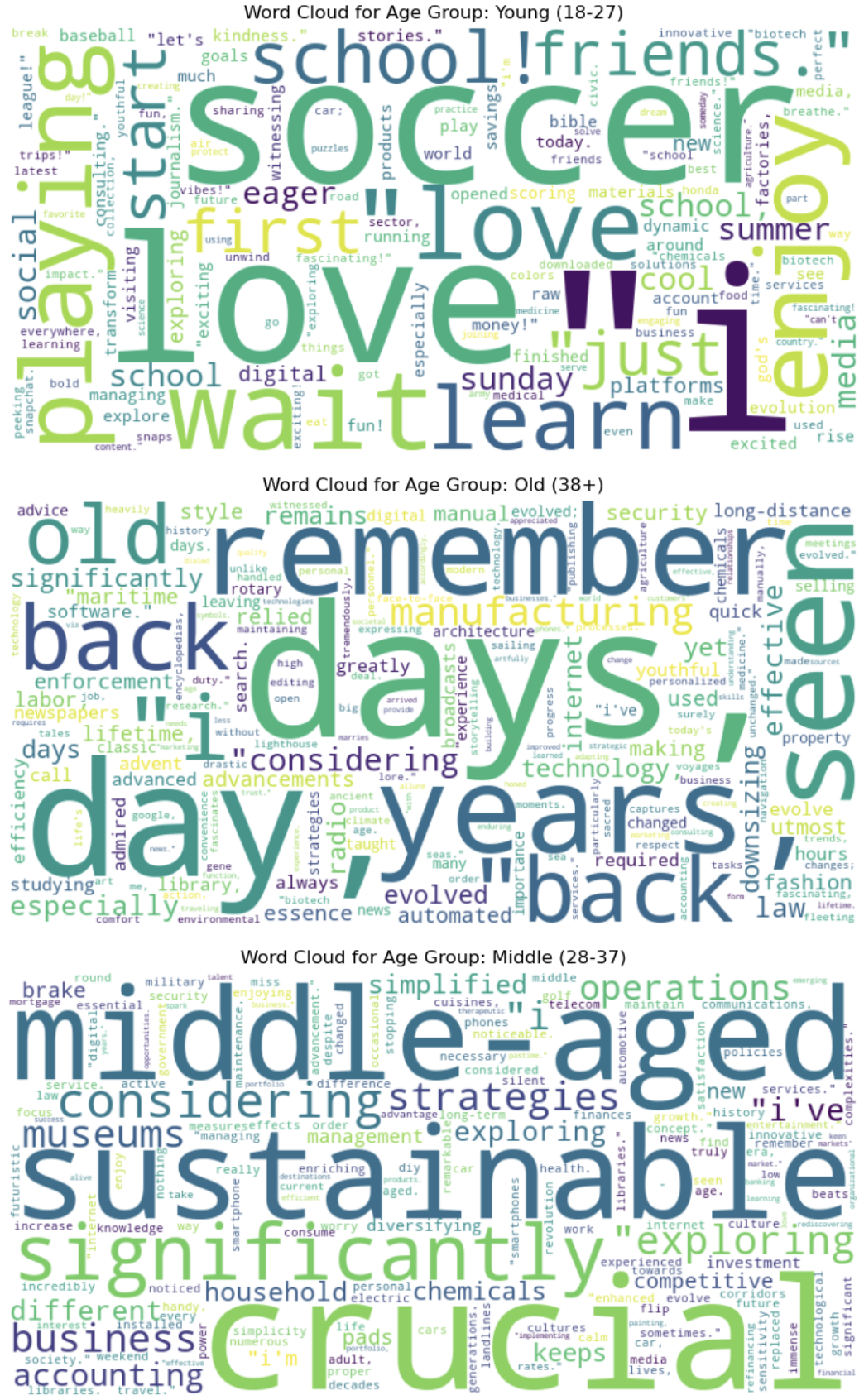}
    \caption{Word CLoud of GPT-4 prompts}
    \label{fig:gpt-wc}
\end{figure}

This runs to create a DataFrame with 1000 samples, each consisting of a piece of generated text, an age, an age group, and a topic. As the function was randomized it allowed me to keep the dataset IID. The word-cloud of the generated text categorized by age group can be seen in Fig \ref{fig:gpt-wc}.

\section{Results}

\subsection{Comparisons with blog text and GPT-4 generated data}
\begin{figure}[htbp]
    \centering
    \includegraphics[width=1\linewidth]{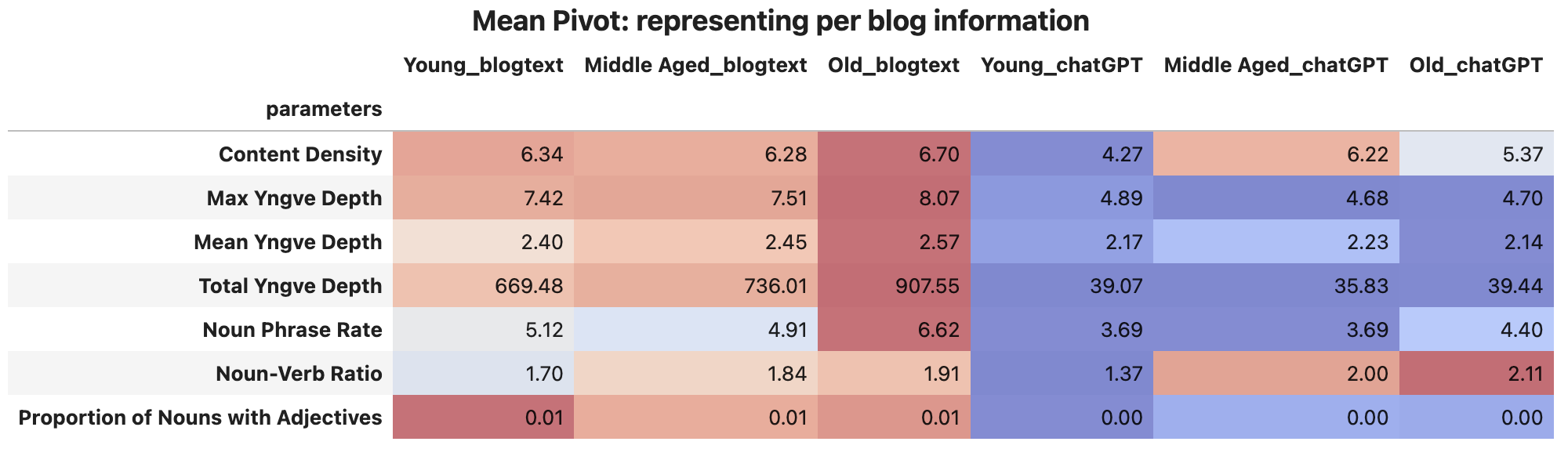}
    \caption{Syntactic Feature Comparison: GPT-4 vs BlogText on the balanced dataset (about 51k rows). Represented as heatmap at row level }
    \label{fig:synt-comp}
\end{figure}

\begin{figure}[htbp]
    \centering
    \includegraphics[width=1\linewidth]{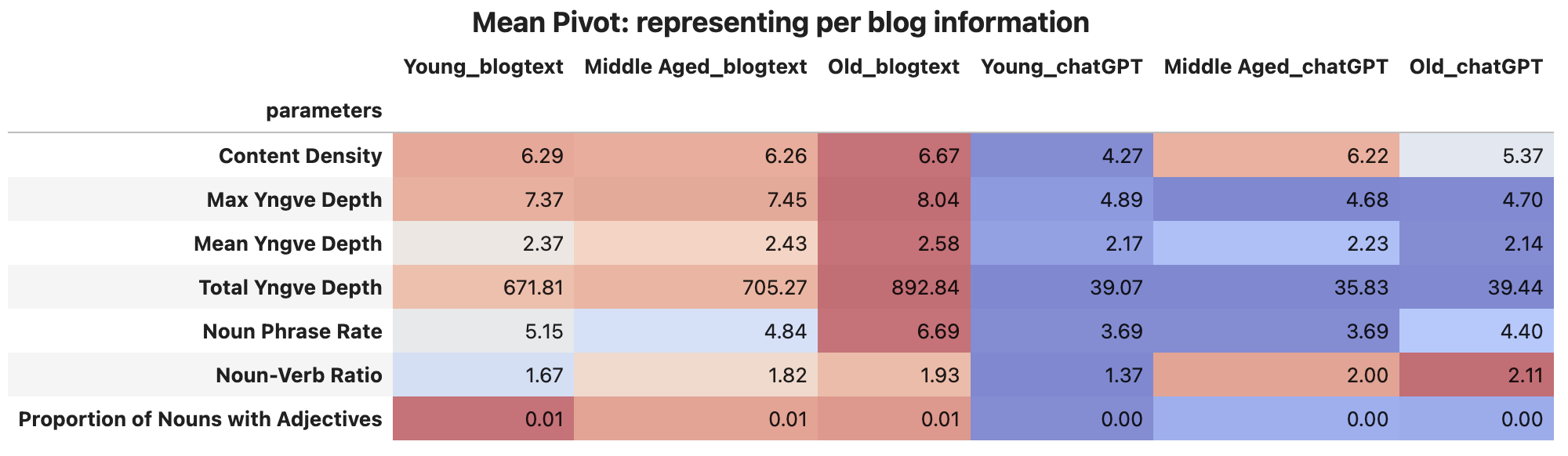}
    \caption{Syntactic Feature Comparison: GPT-4 vs BlogText on the full dataset (about 450k rows)}
    \label{fig:synt-comp-full}
\end{figure}

Fig. \ref{fig:synt-comp} and \ref{fig:synt-comp-full}, illustrate the comparative heatmap of the parsing between different age groups for blog text and GPT-4 text. On careful observation, trends can be seen in the case of blog text which uses a much larger dataset, in these specific metrics depicting the complexity of sentences. It increases with age group increase, while part of speech content majorly remains the same. As the validation dataset from GPT-4 is fairly smaller, exact trends do not replicate but broad changes in sentence complexity can also be seen across age groups. I have picked the key metrics where visible differences could be observed as a trend. The figures \ref{fig:synt-comp} and \ref{fig:synt-comp-full} use randomized and balanced datasets and full datasets for the syntactic parsing table respectively. This comparison gives an encouraging push to the hypothesis where we claim that the experience in language knowledge gives the users confidence in writing complex sentences, being reflected in metrics such as Yngve depth, which is based on dependency parsing.\cite{b9}

\subsection{Forecasting accuracy on new text generated by GPT-4}
\begin{figure}[htbp]
    \centering
    \includegraphics[width=1\linewidth]{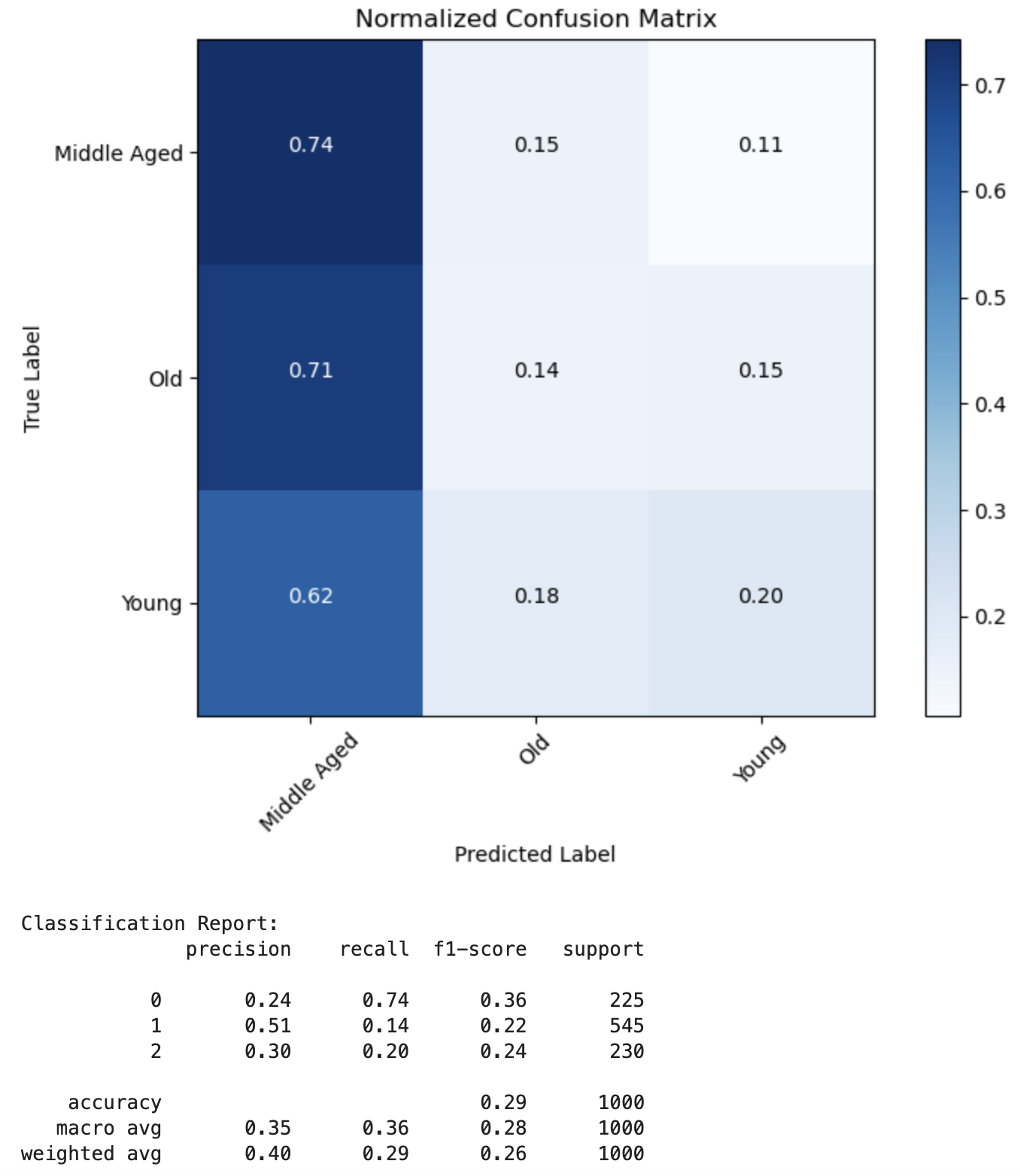}
    \caption{Running Model on GPT-4 data}
    \label{fig:ann_result_gpt}
\end{figure}

Observing the forecasting bit was not so encouraging as when the dataset was trained, the accuracy of the model was low, but when tested on new text from GPT-4, it went down further low. Suggesting either the model in GPT-4 might not be trained on the language evolution of humans, or the dataset from blogger.com is not conclusive enough. (fig.\ref{fig:ann_result_gpt})

\begin{figure}[htbp]
    \centering
    \includegraphics[width=1\linewidth]{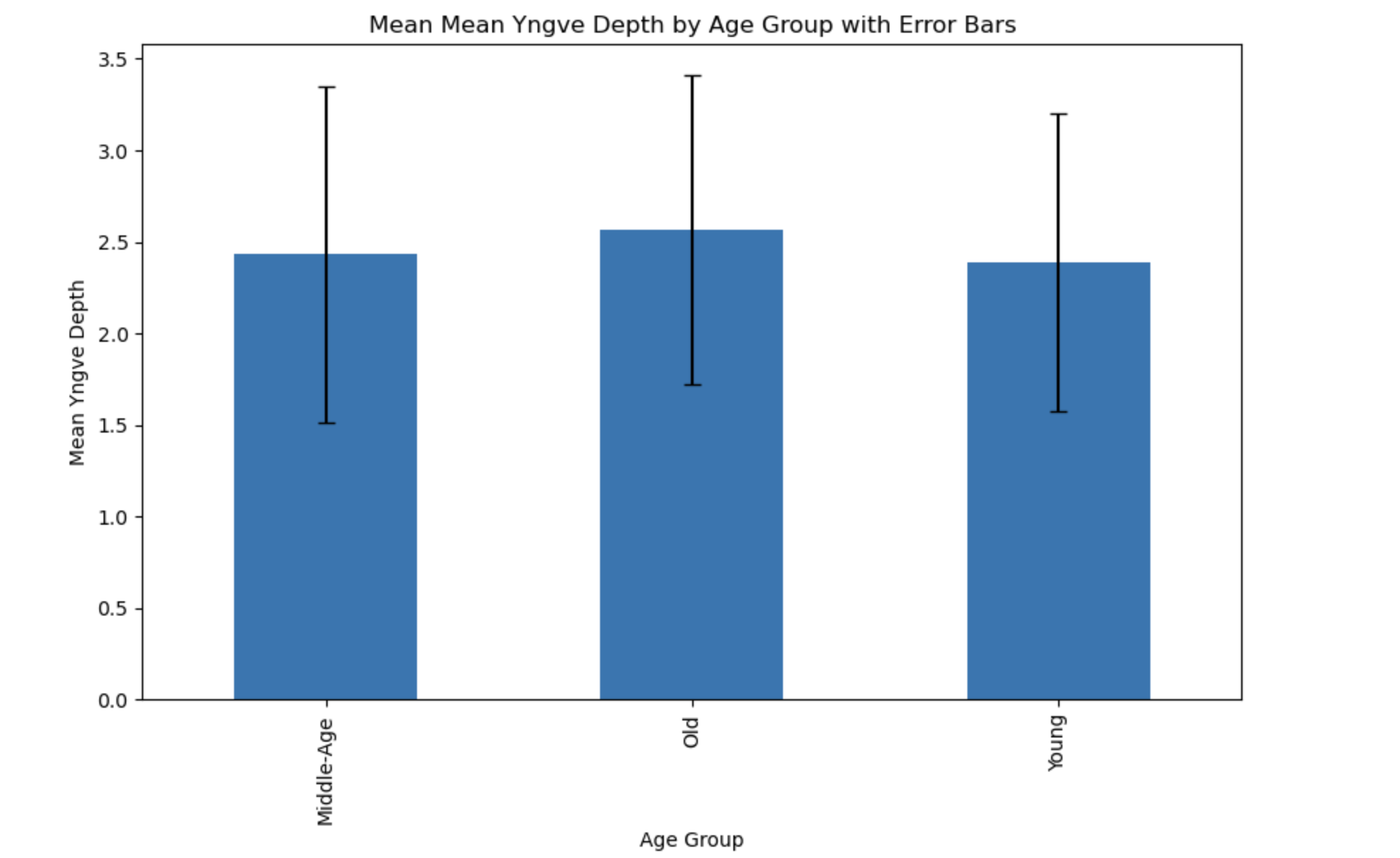}
    \caption{Variance over Bar Graph for Mean Yngve Depth}
    \label{fig:bar_var}
\end{figure}

One probable cause of the forecasting inaccuracy seems to be coming from the variance in metrics used as a feature to forecast, due to non-standard writing in blogs by users. Fig. \ref{fig:bar_var} plots the variance over the bar graph in one of the metrics observed, Yngve depth, where the height of the variance is clearly over 60-70\% of the height of the bar graph itself indicating a high variance. Hence, even though the overall aggregated metrics show a clear trend, forecasting would still be a challenge considering the range of variance. Hence, this also calls for different datasets for training for forecasting the age group based on text, although the difference is majorly known at a higher level based on parsing results.

\subsection{Issues Encountered}
The research journey brought forth several challenges, contributing to a nuanced understanding of the complexities inherent in the study. The primary hurdle emerged in the accessibility of key referenced research papers, which were archived with restricted access. Consequently, reliance on publicly available data became a formidable challenge.

\begin{itemize}
\item Extraction of data revealed the inclusion of authors under 18 years, necessitating their removal to ensure ethical analysis.
\item Notably, the data exhibited skewness toward the young age group, introducing a potential bias in the findings.\\
\item The language model, GPT-4, demonstrated limitations in handling age or maturity-driven linguistic changes, impeding the generation of nuanced and contextually appropriate comparisons.\cite{b5}\
\item Forecasting is not encouraging, as despite deploying a multi-model ensemble (which is a well-known technique in multiclass scenarios \cite{b14}\cite{b15}\cite{b16}) with 5 models stacked in 1st layer and 1 ensemble in the second (logistic regression, random\_forest, svc, gradient\_boosting, mlp\_classifier and over that ran XGboost taking model outputs as features), accuracy achieved was only about 40\%, while on new text from GPT-4 it reduced to about 30\%, with best recall of 74\% in middle-aged class.
\end{itemize}

\subsection{Notes for Future Work}
The encountered challenges served as valuable lessons, providing insights that can inform and refine future research endeavors.

\begin{itemize}
\item Guided by the analysis, patterns, and trends can be identified that may warrant further investigation or hypothesis testing. For instance, exploration could be undertaken to discern whether certain syntactic features correlate with specific topics or sentiments within each age group.

\item To address the temporal gap in the data, a recommendation for future work involves training the model on the latest data, incorporating age\_group and text variables.
\item The imperative of sourcing diverse validation data from multiple channels is emphasized, ensuring a comprehensive and representative dataset.
\item Exploring advanced methodologies, such as an ensemble of artificial neural networks or alternative machine learning techniques, is suggested to enhance the model's accuracy.
\item Beyond the immediate scope, the envisioned future applications of this model extend to the creation of user-focused content in education and fields that demand psychologically inspired communication. This potential highlights the broader societal impact of refining and advancing the current research paradigm. The practical application of the model would still need a further detailed discussion, while the findings show a clear trend. 
\end{itemize}

\subsection{Conclusion}
In this study, our analysis of blogger.com data revealed how writing styles vary with age. This insight is valuable for understanding digital communication in linguistics. It shows how the language used online can change depending on the user's age, which is important for researchers studying language evolution in the digital era.

For communication, these findings help professionals like marketers and content creators. They can use this knowledge to better connect with different age groups online, making their messages more effective.

However, another finding is that AI models like GPT-4 aren't yet very good at matching language styles to specific age groups, specifically complexity across age groups. It often made mistakes when trying to mimic how different ages write as it did not resemble consistently the data from real users. This suggests that more work is needed to improve AI in understanding and replicating human language styles evolved with demographics.

Overall, while our study focuses on a specific set of data, it offers broader insights. It helps us understand language use in the digital world and points out where AI can improve. This can guide future research in linguistics, communication, and AI development.

\section*{Acknowledgment}
I would like to express my sincere gratitude to Professor Alm for her invaluable guidance and unwavering support throughout the entire duration of this research term project. Professor Alm played a crucial role in shaping the trajectory of this work, providing insightful feedback and encouragement at every step.

\vspace{12pt}
\end{document}